\documentclass[suppldata]{interact}

\usepackage{amsmath}
\usepackage{amssymb}
\usepackage{amsthm}
\usepackage{booktabs}
\usepackage{paralist}
\usepackage{hyperref}
\usepackage{xurl}
\usepackage{graphicx}
\graphicspath{{images/}}
\usepackage{wrapfig}

\usepackage{csquotes}
\usepackage{pdfpages}

\usepackage{hyperref} 
\usepackage{epstopdf}
\usepackage{subfigure}

\usepackage{natbib}
\bibpunct[, ]{(}{)}{;}{a}{}{,}
\renewcommand\bibfont{\fontsize{10}{12}\selectfont}

\theoremstyle{plain}

\theoremstyle{definition}

\theoremstyle{remark}

\newcommand{\bignorm}[1]{\big\lVert #1 \big\rVert}

\newcommand{\R}{\mathbb{R}}
\newcommand{\p}{\mathbf{p}}
\newcommand{\la}{\mathbf{l}}

\renewcommand{\l}{\left}

\renewcommand{\r}{\right}

\begin{document}

\title{Generative Design of Ship Propellers using Conditional Flow Matching}

\author{
\name{Patrick Krüger\textsuperscript{a}\thanks{Contact Email: krueger@math.tu-berlin.de}, Rafael Diaz\textsuperscript{b}, Simon Hauschulz\textsuperscript{b}, Stefan Harries\textsuperscript{b}, Hanno Gottschalk\textsuperscript{a}}
\affil{\textsuperscript{a}Institute of Mathematics, TU Berlin, Straße des 17. Juni 135,
10623 Berlin, Germany; \\\textsuperscript{b}FRIENDSHIP SYSTEMS AG, Benzstraße 2, 14482 Potsdam, Germany}
}

\maketitle
\begin{abstract}
In this paper, we explore the use of generative artificial intelligence (GenAI) for ship propeller design. While traditional forward machine learning models predict the performance of mechanical components based on given design parameters, GenAI models aim to generate designs that achieve specified performance targets. In particular, we employ conditional flow matching to establish a bidirectional mapping between design parameters and simulated noise that is conditioned on performance labels. This approach enables the generation of multiple valid designs corresponding to the same performance targets by sampling over the noise vector.

To support model training, we generate data using a vortex lattice method for numerical simulation and analyze the trade-off between model accuracy and the amount of available data. We further propose data augmentation using pseudo-labels derived from less data-intensive forward surrogate models, which can often improve overall model performance. Finally, we present examples of distinct propeller geometries that exhibit nearly identical performance characteristics, illustrating the versatility and potential of GenAI in engineering design.
\end{abstract}

\begin{keywords}
Ship propeller design, generative AI, NACA parametric description, vortex lattice method, data requirement, data augmentation, case study
\end{keywords}

\section{Introduction}

\subsection{Advancements in Ship Propeller Design}
The design workflow of ship propellers has changed significantly since the appearance of simulation-driven design (SDD) techniques in marine engineering. This approach leverages advanced computational tools such as computational fluid dynamics (CFD) and optimization algorithms to optimize a propeller geometry and its simulated performance. Geometrical properties are, e.g., the diameter, the number of blades, and the pitch distribution of a blade. Desired propeller performance can be described by, e.g., thrust, torque, efficiency, and cavitation criteria. SDD not only accelerates the development process but also enables engineers to handle more complex problems. It further minimizes the need for physical prototypes and replaces traditional engineering workflows, which typically rely on experience and trial-and-error approaches.\\
In the 2000s and 2010s, the high computational cost and time associated with flow simulations led to the development of surrogate models to mimic the behavior of these simulations but with a significantly reduced computational effort. These data-driven approaches rely on statistical models, which require enough data from previous or offline simulations to approximate the simulation outcomes reasonably well. Examples of these models are response surface methods, kriging, and neural networks, to name a few. The fast computation time of surrogate models also makes them ideal for tasks like design space exploration and sensitivity analysis, where numerous evaluations are required.\\
In the past decade machine learning has made notable progress with the development of generative models, which are designed to learn the underlying structure of their training data, so that they can generate new similar data. This led to tools such as DALL-E \cite{DallE} for image generation or ChatGPT \cite{ChatGPT} for text generation, which became popular in recent years. In the context of engineering, generative models may be used to explore a wide range of design options efficiently. Especially, the inverse design approach presents promising applications because instead of defining a design and evaluating its target performances with a simulation or surrogate model, one may first set desired performances and let the generative model make suggestions for suitable designs that achieve them.

\subsection{Engineering and Generative AI}

Generative artificial intelligence (GenAI) is influencing almost every scientific discipline. It is not a daring prediction that GenAI in engineering sciences will be disruptive, since the creation of new designs that meet certain performance requirements is a creative act. As such, it should be close to what GenAI models are made for.  

Data availability, however, stands in the way of building successful GenAI models for engineering. Data that relates to design is often kept private in order to distinguish oneself from the competition. Therefore Large Language Models have not yet been trained at scale on engineering data. This gap, as we show in this article, can however be filled by smaller, domain specific generative models.

This leads to the question, how scientists and companies with very specific domain knowledge can become data driven companies utilizing GenAI. Often, commercial or public domain expert organizations possess simulation tools to produce large amounts of domain specific data. To become data-driven therefore first of all means to decouple data production from specific design tasks and produce such data simply because of its inherent value. GenAI models trained on such data pools should then extract the useful information and be capable to provide design proposals instantaneously.

The good news is that the principles underlying GenAI do not necessarily require a huge general purpose model. Small GenAI models can be trained to encode the relationship between requirements (performance labels) and design parameters. Unlike surrogate models that have been used in engineering for a long time, GenAI models do not (only) learn to map design parameters to performance labels, but also solve the 'inverse' question to generate design parameters matching predefined labels.  

This work proposes flow based models (\cite{AffCoupling,chan2023lu,NeuralODEs,FlowMatchLipman})  to generate design proposals that meet specific performance requirements. First introduced as Normalizing Flows (\cite{rezende2015variational}), these models transform a tractable distribution - typically a Gaussian - into a complex data distribution of interest, e.g., design parameters given certain requirements. In Normalizing Flows and their subsequent iterations, this transformation is built by learning a series of explicitly invertible mappings. This probabilistic framework enables the generation of diverse design candidates that are statistically consistent with prescribed performance objectives. For instance, flow based Invertible Neural Networks (INN)  (\cite{ardizzone2018analyzing}) were employed in the design of gas turbine combustors (\cite{GasTurb}), airfoil blades (\cite{jia2025inverse}) and thermophotovolatic emitters (\cite{yang2023normalizing}).

In the present study, we extend these generative modeling approaches to the domain of ship propeller design. Whereas previous works have employed INN (\cite{ardizzone2018analyzing}) to construct explicit bijective mappings between design parameters and performance labels, we adopt the more recent Flow Matching algorithm (\cite{FlowMatchLipman}). Flow Matching belongs to the family of continuous normalizing flows (\cite{NeuralODEs}) and learns a time-dependent vector field that transports samples from a base distribution toward the empirical data distribution. Unlike e.g. INNs, it does not impose strict pointwise invertibility between parameters and labels but instead models the conditional probability flow underlying the data generation process by a learned ordinary differential equation. The proposed model is trained on simulation data derived from a parametric representation of ship propellers. Additionally, we introduce an augmentation strategy incorporating a traditional forward surrogate model, which demonstrably improves training stability and predictive fidelity for small- to medium-sized datasets.

\subsection{Structure of this Paper}

We start in \autoref{chap:PropData} with the definition of the dataset consisting of the parameters for the geometry and the labels for the performance of a propeller. We then briefly explain the numerical simulation used to derive these labels from the parameters. Here, CAESES by Friendship Systems facilitated the definition and illustration of parameterized propeller geometries and the implementation of flow simulations, providing the data for the machine learning model. In \autoref{chap:MathFound} we give a mathematical introduction by presenting key concepts which are necessary to understand the Conditional Flow Matching (CFM)  framework. After describing the model architecture and the training procedure, we use the dataset to train and validate a generative model as described in \autoref{chap:TrainVal}. In \autoref{chap:Results} we analyze the performance of the CFM model and also conduct further studies on the generative capabilities of the model, covering both accuracy and diversity of generated designs. Here, we also conduct an additional study on model performance after data augmentation with surrogate models and second while reducing the originally available amount of training data. Finally, in \autoref{chap:Summary} we conclude our results and provide prospects for further development in this field.

\section{Related Works}
The review \cite{GenAIEngineering} provides an overview of the applications of Generative Adversarial Networks (GAN) (\cite{GAN}), Variational Autoencoders (VAE) (\cite{VAE}), and Reinforcement Learning (\cite{RL}) models in structural optimization, material design, and shape synthesis, while also discussing the challenges arising in the design procedure when using generative models. Another review focusing on the inverse design of materials is given by \cite{InvDesignMaterial}. For the aerodynamic design of an airfoil, such as propeller blades or aircraft wings, \cite{AirfoilVAE} and \cite{AirfoilCGAN} discuss an inverse design strategy to optimize aerodynamic performances using VAE and conditional GAN, respectively. In the marine engineering domain, \cite{GenAIShipDesign} used Gaussian Mixture Models (\cite{GMM}) to learn the underlying patterns of a ship hull dataset and generate new ship hull designs, which were then filtered by a target label, that is, the total resistance of the hull. Another approach, which incorporates target labels directly into the generative model, is described in \cite{InvProbINN}, which uses invertible neural networks consisting of affine coupling layers (\cite{AffCoupling}). \cite{GasTurb} applied this approach to the design of a gas turbine combustor in order to generate designs that meet the specified target labels.

\section{Propeller Dataset Definition}
\label{chap:PropData}
This Section covers the creation of the propeller dataset used to train the generative model. Points in the dataset are given by fully-parameterized propeller geometries defined by design variables described in Section \ref{sect:PropParam} and associated performance labels discussed in sub-section \ref{sect:PropLabel}. In sub-section \ref{sect:FlowSim}, the derivation of performance labels by a CFD simulation based on the vortex lattice method (VLM) is described. Finally, Section \ref{sect:DataGenAna} covers the generation of the dataset and highlights inherent connections of design variables and labels.\\
\ \\
The theory of propeller geometry and VLM can be found in \cite{PropTheory} and \cite{VLM}, respectively.
To get an introduction to propeller design, we refer to the CAESES documentation\footnote{CAESES Documentation: \url{https://docs.caeses.com/docs/propeller/fundamentals/introduction}}.

\subsection{Fully-Parametric Propeller Model}
\label{sect:PropParam}
For the dataset, a fully-parametric propeller geometry was set up using the CAESES Propeller Design Workflow\footnote{CAESES Documentation: \url{https://docs.caeses.com/docs/propeller/propeller-workflow}} which leads the designer in a step-by-step generation process to a robust and variable model. 
The CAESES Meta Surface functionality\footnote{CAESES Documentation: \url{https://docs.caeses.com/docs/geometry-modeling/surfaces\#meta-surface-1}} allows to vary the propeller shape by using design variables to change the radial distributions of the usual parameters, e.g., chord, pitch, camber. For this study, only a small subset of possible design variables was introduced and varied. Each propeller geometry is defined by a vector $\mathbf{p}$ of six independent design variables:

\begin{equation}\label{eq:design_variables}
    \mathbf{p}=(N_b,P,W_{rp},W_c, W_{rc},C).
\end{equation}

\begin{figure}
\begin{minipage}{.6\textwidth}
    \centering
    \includegraphics[width=\linewidth]{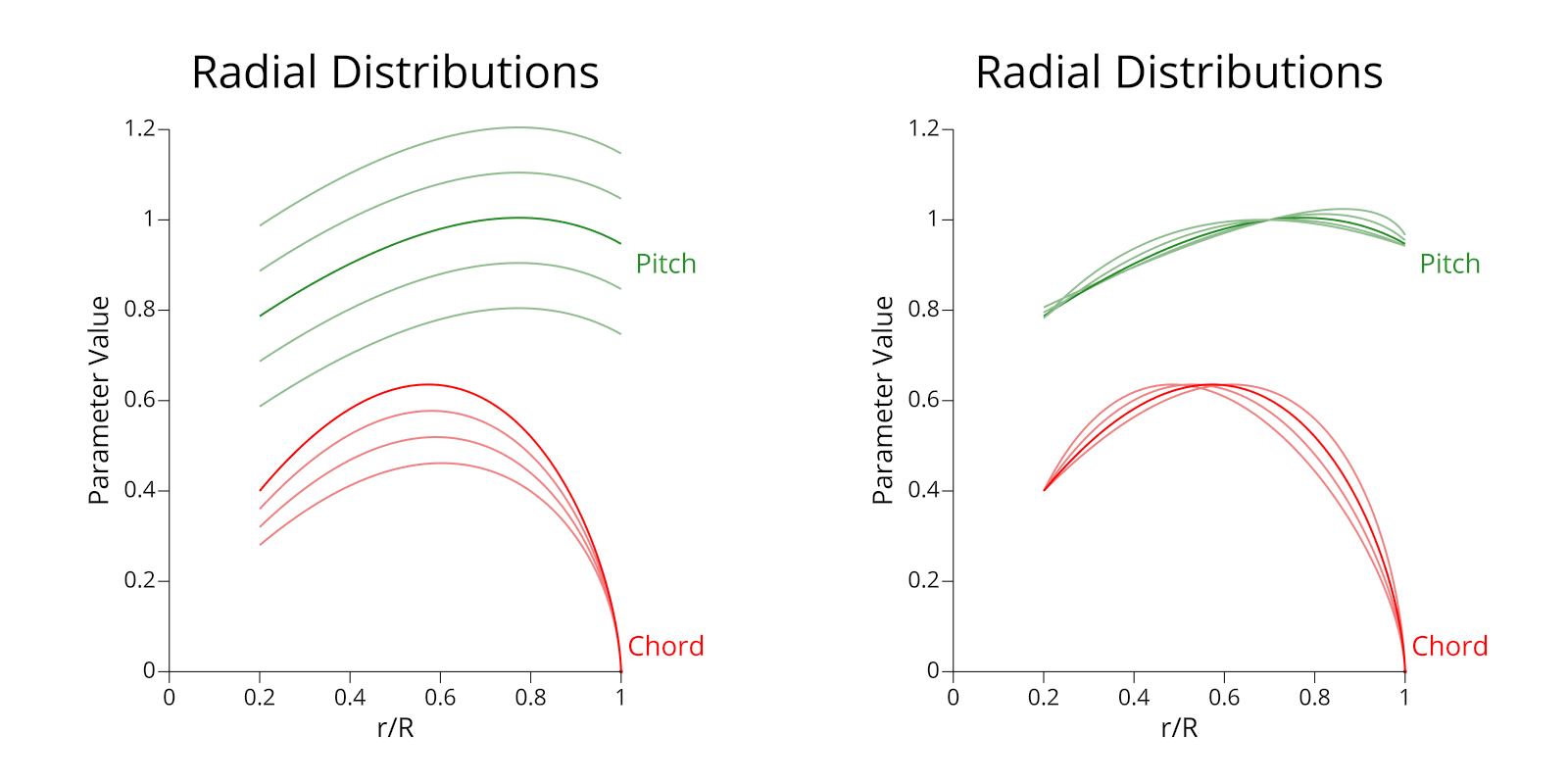}
    \caption{Variation of Chord and Pitch with $P$ and $W_c$ (left) and $W_{rc}$ and $W_{rp}$ (right)}
    \label{fig:radialDist_variation}
\end{minipage}
\hfill
\begin{minipage}{.35\textwidth}
    \centering
    \includegraphics[width=\linewidth]{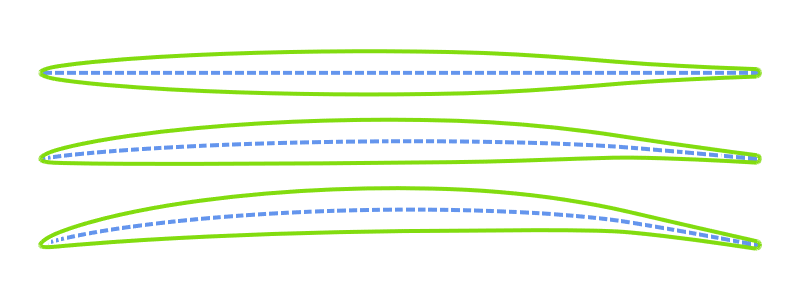}
    \caption{Camber variation of NACA 66 profile}
    \label{fig:profileCamber}
\end{minipage}
\end{figure}

The design variables listed in Eq. \ref{eq:design_variables} are the number of blades $N_b$, an offset to a pre-selected pitch distribution $P$, a weight to shift the maximum pitch position $W_{rp}$, the weights $W_c$ and $W_{rc}$ to shift the maximum chord value and its position, and the camber value $C$.\\
An overview of the design variables of the propeller dataset and their ranges is given in \autoref{tab:prop-params}. \autoref{fig:radialDist_variation} visualizes the influence of $P$, $W_{rp}$, $W_c$ and $W_{rc}$ on the radial distributions of chord and pitch.

\begin{table}[h]
    \tbl{Independent parameters of the propeller dataset used as design variables}
  {\begin{tabular}{c l c}
    \toprule
    \textbf{Parameter} & \multicolumn{1}{c}{\textbf{Description}} & \textbf{Range} \\ \midrule
    $N_b$       & Number of blades & $\{2,3,4,5\}$ \\
    $P$         & Nominal pitch; shifts pitch distribution curve accordingly & $[0.5,1.5]$ \\
    $W_{rp}$    & Weight to control the radial position of the maximal pitch  & $[0.5,0.9]$ \\
    $W_c$       & Weight to control the maximal chord & $[0.5,1]$ \\
    $W_{rc}$    & Weight to control the radial position of the maximal chord & $[0.5,0.8]$ \\
    $C$         & Constant camber along the radius, except for the blade tip & $[0,0.05]$ \\ \bottomrule
  \end{tabular}}
  
  \label{tab:prop-params}
\end{table}

\begin{figure}[h]
\centering
\begin{minipage}{.45\textwidth}
  \centering
  \includegraphics[width=0.65\linewidth]{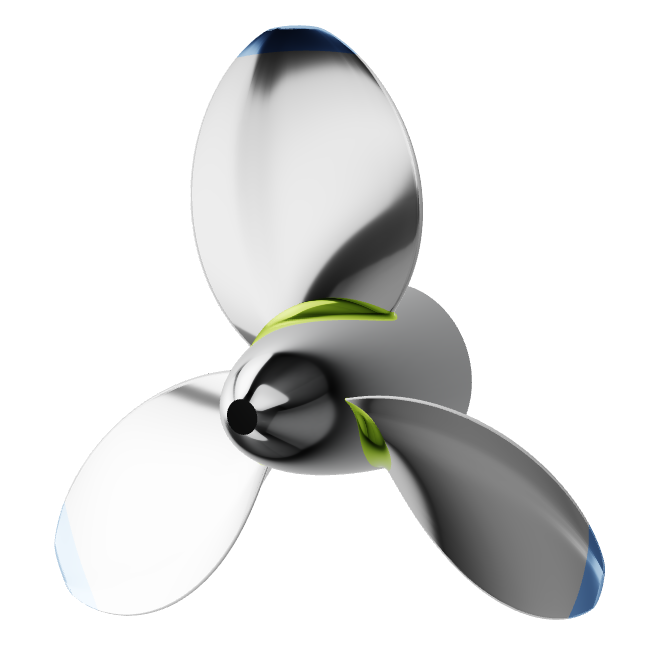}
  \caption{An exemplary propeller created with CAESES}
  \label{fig:prop}
\end{minipage}
\hfill
\begin{minipage}{.45\textwidth}
  \centering
  \includegraphics[width=0.65\linewidth]{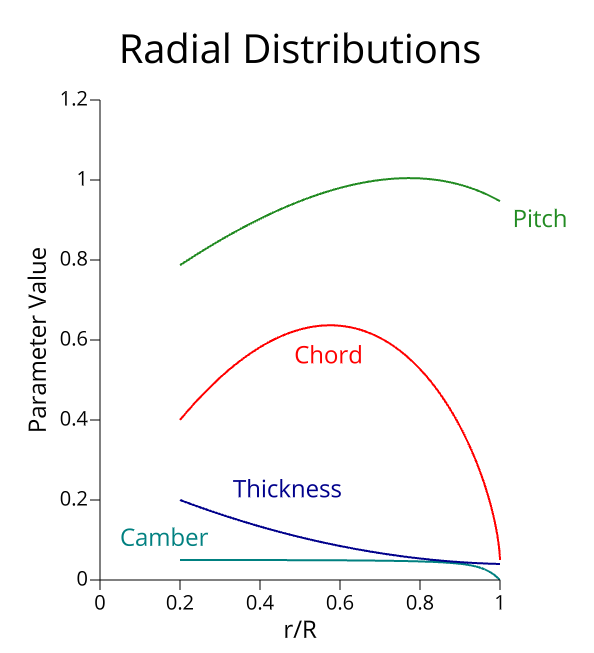}
  \caption{Radial distribution functions of the propeller shown in \autoref{fig:prop}}
  \label{fig:prop-params}
\end{minipage}
\end{figure}

The radial distributions for thickness, rake and skew were not varied in this study since its aim was to investigate the applicability of the methods rather than the generation of data already applicable for practical design tasks. Skew and rake, which have a secondary effect on the propeller performance compared to the chosen design variables and their corresponding radial distributions, were set to zero along the radius and the thickness along the radius was fixed to a reasonable distribution. The blade surface is generated from NACA 66 mod a-0.8 sections, a profile definition specifically designed for marine propeller applications (\cite{Naca66}). \autoref{fig:profileCamber} illustrates the variation of the camber value on one section. \autoref{fig:prop} shows an exemplary propeller created with the design workflow and the radial distributions of this design are shown in \autoref{fig:prop-params}.

\subsection{Performance Labels}
\label{sect:PropLabel}
The performance labels were chosen on the basis of a typical propeller design process, where the advance velocity $V_A$, the propeller rpm $n$, the required thrust $T$ and the propeller diameter $D$ are given by the use case and the design of the ship hull. This leads to the desired operating condition, i.e., the advance ratio $J:=\frac{V_A}{nD}$ and the thrust coefficient $k_T$ at this operating condition. Ideally, the final propeller design has the lowest torque coefficient $k_Q$ or, equivalently, the highest efficiency $\eta := \frac{J}{2\pi} \frac{k_T}{k_Q}$ at the desired $J$.\\
Note, the workflow presented here is not about optimizing the propeller design over $\eta$, but rather about generating suitable designs for given performance labels. Thus, for any given propeller design, we define the performance labels
\begin{equation}
\label{eq:labels}
    \mathbf{l}:=\big(\eta^*,J^*,k_T^*\big),
\end{equation}
where $\eta^*$ is the maximal efficiency of the propeller across several operating states, $J^*$ is the operating state for which $\eta^*$ is reached and $k_T^*$ is the thrust coefficient at $J^*$. An overview of the performance labels \eqref{eq:labels} and their value ranges is given by \autoref{tab:prop-labels}. With these labels, it is possible in practice to enter the desired $J$, $k_T$ and a slowly increasing $\eta$ into the generative model, while inspecting and evaluating the suitability of the generated designs.

\begin{table}[h]
  \tbl{Performance labels of the propeller dataset}
  {\begin{tabular}{clc}
    \toprule
    \textbf{Label} & \multicolumn{1}{c}{\textbf{Description}} & \textbf{Range} \\ \midrule
    $\eta^*$       & Maximal propeller efficiency & $[0.454, 0.863]$ \\
    $J^*$    & Operating state for which $\eta^*$ is attained & $[0.335, 1.466]$ \\
    $k_T^*$         & Thrust coefficient at the operating state $J^*$ & $[0.031, 0.192]$ \\ \bottomrule
  \end{tabular}}
  \label{tab:prop-labels}
\end{table}

\subsection{Flow Simulation}
\label{sect:FlowSim}
In order to obtain the performance labels of the propellers, we performed simulations using the OpenProp software\footnote{OpenProp Homepage: https://www.epps.com/openprop} which is based on moderately loaded lifting line theory and is implemented in MATLAB\footnote{MATLAB Software: https://de.mathworks.com/products/matlab.html} Details about the implementation and the theory involved can be found in \cite{OpenProp}. The use of a low-fidelity solver that is less computationally expensive than a RANS solver allowed building a large dataset to test the inverse design approach.

\subsubsection{OpenProp Simulation Procedure}
To run a simulation with OpenProp we first need to specify a range of operating states $\mathcal{J}=\{J_1,\dots,J_n\}$ of the propeller, which we specify by
\begin{equation}
\label{eq:op-states}
    \mathcal{J}:=\{0.25, 0.3, 0.35, \dots, 1.6\}.
\end{equation}
OpenProp models the blade as discrete sections given at normalized radii $\mathcal{R}=\{\frac{r_1}{R},\dots,\frac{r_m}{R}\}$, which we set to be
\begin{equation}
\label{eq:radii}
    \mathcal{R}:=\{0.2, 0.3, 0.4, 0.5, 0.6 ,0.7, 0.8, 0.9, 0.95, 1\}.
\end{equation}
For each radius listed in \eqref{eq:radii}, we provide the section's pitch, chord, camber, and thickness as the respective evaluation of the radial distribution. Finally, for each radius, the section drag coefficient is set to $C_D=0.008$, the axial inflow-to-ship velocity ratio is set to $\frac{V_a}{V_s}=1$, and the tangential inflow-to-ship velocity ratio is set to $\frac{V_t}{V_s}=0$. This corresponds to an open-water set-up, delibaretly accepting that propeller-hull interactions are not accounted for.\\
Based on these inputs, OpenProp uses a VLM to derive a set of nonlinear equations, which then are solved iteratively using Newton's method. In this way, the thrust and torque coefficients are determined for each operating condition.

\subsubsection{Evaluation of an Exemplary Simulation}
\begin{figure}
    \centering
    \includegraphics[width=0.4\linewidth]{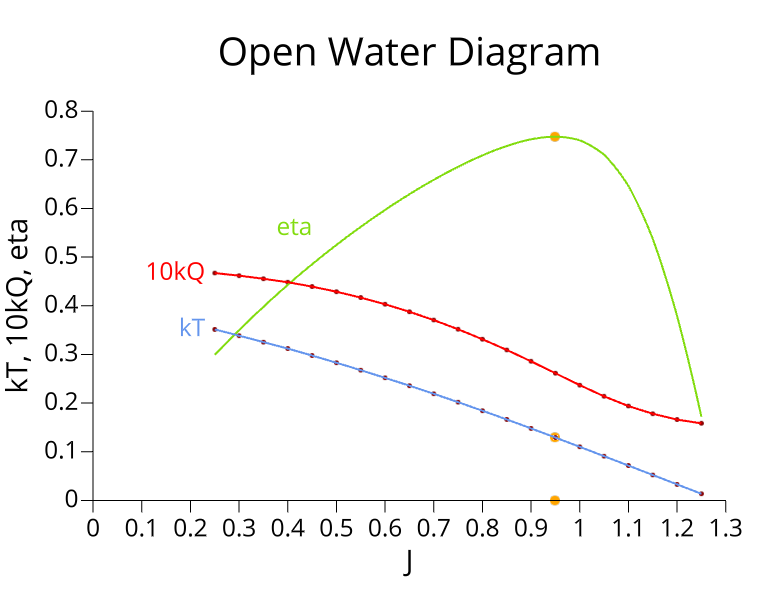}
    \caption{Open Water Diagram of the propeller shown in \autoref{fig:prop}}
    \label{fig:prop-labels}
\end{figure}

To outline the evaluation of the OpenProp simulation, a simulation with the propeller example shown in \autoref{fig:prop} was carried out. From the results, the set $\{{k_T}_1, \dots, {k_T}_n\}$ of thrust coefficients corresponding to $J_1,\dots,J_n$ listed in \eqref{eq:op-states} are interpolated by a piecewise linear function representing the thrust coefficient $k_T$ over $J$. The torque coefficient $k_Q$ over $J$ was derived analogously. These interpolations approximate the real thrust and torque coefficient functions sufficiently well, as its curvature along $J$ is usually small. The propeller efficiency over $J$ is then given by $\eta(J) := \frac{J}{2\pi} \frac{k_T(J)}{k_Q(J)}$.\\
The thrust, torque and efficiency curves are visualized in an open water diagram as shown in \autoref{fig:prop-labels}. The performance labels were then derived from the maximum of the efficiency curve at the point $(J^*, \eta^*)$ and the corresponding thrust coefficient $k_T^*$, which are colored orange in the diagram. These three values then correspond to the performance label $l=\big(\eta^*,J^*,k_T^*\big)$ of the propeller. It should be noted that common practice of choosing the design point just a bit the left of the maximum efficiency is neglected for sake of simplicity. 

\subsection{Dataset Generation and Analysis}
\label{sect:DataGenAna}
\begin{figure}
    \centering
    \includegraphics[width=\linewidth]{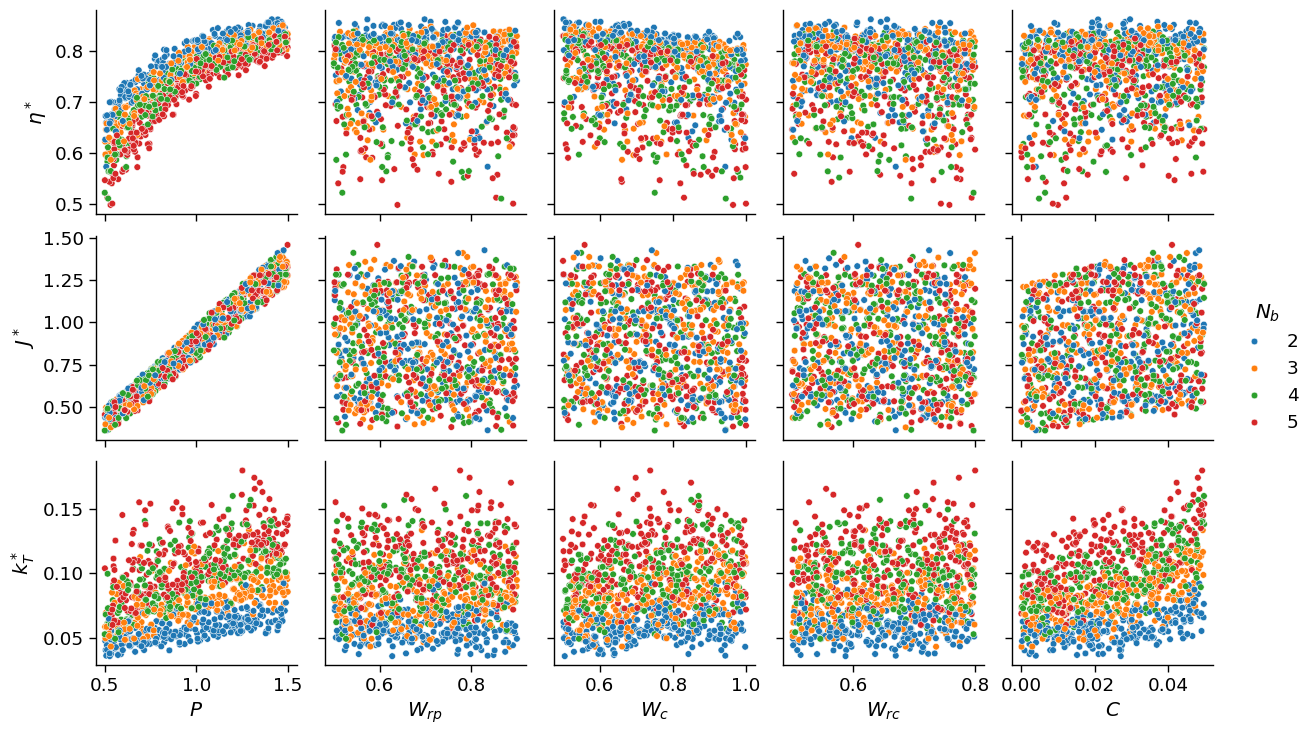}
    \caption{Overview of the test set showing the correlation between the design variables ($x$-axis) and the performance labels ($y$-axis)}
    \label{fig:pairplot}
\end{figure}

With the procedure described above, 3000 independent design variable vectors $\mathbf{p}$ were sampled using Latin hypercube sampling (\cite{LHS}) and the corresponding label vector $\mathbf{l}$ defined in Eq. \eqref{eq:labels} was derived. 2000 of those samples form the training dataset $\mathcal{D}_{\mathrm{Train}}$. The remaining 1000 samples form the test dataset $\mathcal{D}_{\mathrm{Test}}$ used to validate the trained models.\\
An overview of $\mathcal{D}_{\mathrm{Test}}$ is given in \autoref{fig:pairplot}. Of all design variables, the pitch design variable $P$ has the strongest correlation with the three performance labels. Increasing $P$ leads to an increase of $\eta^*$, $J^*$ and $k_T^*$. Furthermore, the number of blades $N_b$ has a large effect on the possible $k_T^*$ and $\eta^*$ values. A propeller with two blades will be able to reach a higher maximum efficiency than a five-bladed one, but is typically not able to produce as much thrust at the respective operating condition.\\
These two trends should be taken into account when evaluating the designs proposed by the inverse design approach.

\section{Generative Modeling Using Conditional FlowMatching}
\label{chap:MathFound}
Conditional generative learning is the approximation of an unknown distribution $q(x\mid y)$ of data samples $x$ (e.g., Images or Text) that depend on conditions $y$ (e.g., image descriptions or text prompts) using a neural network. In this work, the target distribution is $q(\mathbf{p}\mid \la)$ of design parameter vectors $\mathbf{p}$ of the propeller model (sub-section \ref{sect:PropParam}) given specified label vectors $\la$ (sub-section \ref{sect:PropLabel}). The trained generative model can thus propose new design alternatives $\p$ that meet desired performance criteria. Several well known approaches for conditional generative modeling exist, amongst them are GAN (\cite{GAN}), Normalizing Flows (\cite{kingma2018glow}) and Invertible Neural Networks (\cite{ardizzone2018analyzing}). In this work, the novel Conditional Flow Matching (CFM) method is used as described in \cite{CondFlowMatchTong,FlowMatchLipman} to model the conditional distribution. CFM itself is an approach to efficiently train Neural Ordinary Differential Equations \cite{NeuralODEs}, which will be covered in section \ref{sect:NODE}. An overview of Conditional Flow Matching is given in Section \ref{sect:FlowMatch}.

\subsection{Neural Ordinary Differential Equations}
\label{sect:NODE}
Neural Ordinary Differential Equations (NODE, \cite{NeuralODEs}) describe the continuous change of a time-dependent state via an ODE: 
\begin{align}
    \label{eq:ode}
    \frac{d}{dt} x(t) &= v_t^\theta(x(t)), \\
    \label{eq:init-cond}
    x(s) &= x_s \in \R^d,
\end{align}
in Equation \eqref{eq:ode}, $x(t)$ is the state at time $t$ and $v_t^\theta\colon\R^d\to\R^d$ is the time-dependent vector field that defines the ODE and is described by a neural network with network parameters $\theta$. Equation \eqref{eq:init-cond} describes an initial condition imposed on the ODE. 
One now defines the \textit{flow} $\Phi_{s,t}\colon\R^d\to\R^d$ of the vector field $v_t^\theta$ as $\Phi_{s,t}(x_s) := x(t)$ for some $s,t\in[0,1]$, where $x$ is the solution to the ODE \eqref{eq:ode} given the initial condition \eqref{eq:init-cond}. $\Phi_{s,t}$ thus maps the state $x_s$ given at time $s$ to the state at some time $t$ according to the dynamics described by the vector field $v_t^\theta$.
In the context of generative learning, Neural ODEs are used as so-called \textit{continuous normalizing flows} by defining input data $x(0)=x_0$ that follows a well known and easy to sample source distribution $q_0$ as an initial value problem to \eqref{eq:ode}. In most cases, the source distribution is simply chosen to be a multivariate Gaussian. Furthermore, samples from the unknown target distribution $q$ that is to be learned are defined as solutions $x(1)=x_1$ at time $t=1$. Once $v_t^{\theta}$ is trained, the flow $\Phi_t:=\Phi_{0,t}$ describes a continuous transformation of samples $x_0$ from the simple distribution $q_0$ into samples from the target distribution $q_1=q$. Moreover, for $t\in[0,1]$ the pushforward measure $p_t := \Phi_{t\,*}\,q_0$ describes a continuous transformation of the source distribution $q_0$ into the target distribution $q_1$ that is called the \textit{probability path} generated by the vector field $v_t^\theta$.

\subsection{Conditional Flow Matching}
\label{sect:FlowMatch}
Training neural ODEs as described in \cite{NeuralODEs} is  not scalable as in each training step, the ODE needs to be solved multiple times by a numerical ODE Solver. 
Conditional Flow Matching (\cite{FlowMatchLipman}, \cite{CondFlowMatchTong}) aims to remove the necessity of an ODE solver during training by directly regressing the vector field $v_t$ (Eq. \eqref{eq:ode}) against target vector fields that are constructed on a per-sample basis:

\begin{equation}
\label{eq:cfm-loss}
    \mathcal{L}_{CFM}(\theta) := 
    \mathbb{E}_{z\sim\gamma,\, t\sim\mathcal{U}(0,1),\, x\sim p_t(\cdot|z)}
    \l[\bignorm{v_t^\theta(x)-u_t(x|z)}^2\r].
\end{equation}
In Equation \eqref{eq:cfm-loss}, $p_t(x\mid z)$ is a conditional probability path that is generated by the flow of the conditional vector field $u_t(x\mid z)$. Both $p_t$ and $u_t$ depend on a conditioning variable $z\sim q_z$ that is known and easy to sample. For CFM Models used in this work, $q_z$ represents the independent sampling from source and target distribution, i.e.,  $z=(x_0,x_1)$ and $q_z=q_0q_1$. Given such a sample, $p_t(x\mid z)$ is defined as the linear interpolation between $x_0$ and $x_1$:
\begin{equation}\label{eq:pt}
    p_t(x\mid z)=tx_1+(1-t)x_0.
\end{equation}
It can be proven (\cite{CondFlowMatchTong}, Theorem 2.1) that, with $p_t$ given as in Eq. \eqref{eq:pt}, the conditional vector field $u_t(x\mid z)$ simplifies to
\begin{equation}
    u_t(x\mid z)=x_1-x_0.
\end{equation}
With $z,p_t$ and $u_t$ as above, the calculation of $\mathcal{L}_{\mathrm{CFM}}$ becomes tractable. Upon reaching zero loss, the trained vector field $v_t^\theta(x)$ generates the desired probability path from the source distribution to the target distribution, i.e., $q_1(x)=q(x)$. Thus, new samples of $q$ can be generated by drawing samples $x_0$ from the simple source distribution $q_0$ and transporting them along the flow $\Phi_t$ of $v_t^\theta(x)$.

\section{Training Setup, Surrogate Models and Data Augmentation}
\label{chap:TrainVal}
In this chapter, we describe our experimental setup used to train and test the generative models. The training procedure and configuration of the CFM models used for the experiments in sub-section \ref{chap:Results} are described in sub-section \ref{sect:FMModel}. For a fast validation of different CFM model configurations, additional surrogate models were employed as described in sub-section \ref{sect:SUModel}. Computational time and resources required for the training of both surrogate and CFM model is briefly covered in Section \ref{sect:TimeRes}.

\subsection{CFM Model}
\label{sect:FMModel}
The CFM model defining the vector field $v_t^{\theta}$ in Eq. \ref{eq:cfm-loss} was a simple feed forward neural network implemented using the \texttt{PyTorch} framework. Calculation of probability paths, target vector fields and flows (see sub-section \ref{chap:MathFound}) was realized within the \texttt{TorchCFM} library (\cite{TorchCFM}). The CFM model consists of $8$ hidden layers of $500$ neurons each. The activation function applied to each hidden layer was the rectified linear unit (ReLU, \cite{fukushima1975cognitron}). No activation was applied to the output layer. Using an Adams optimizer, training ran over $10000$ epochs with a batch size of $500$ and an initial learning rate of 0.001, which was reduced by a factor of 10 after $5000$ epochs. 

\subsection{Surrogate Models}
\label{sect:SUModel}

\begin{figure}
    \centering
    \includegraphics[width=\linewidth]{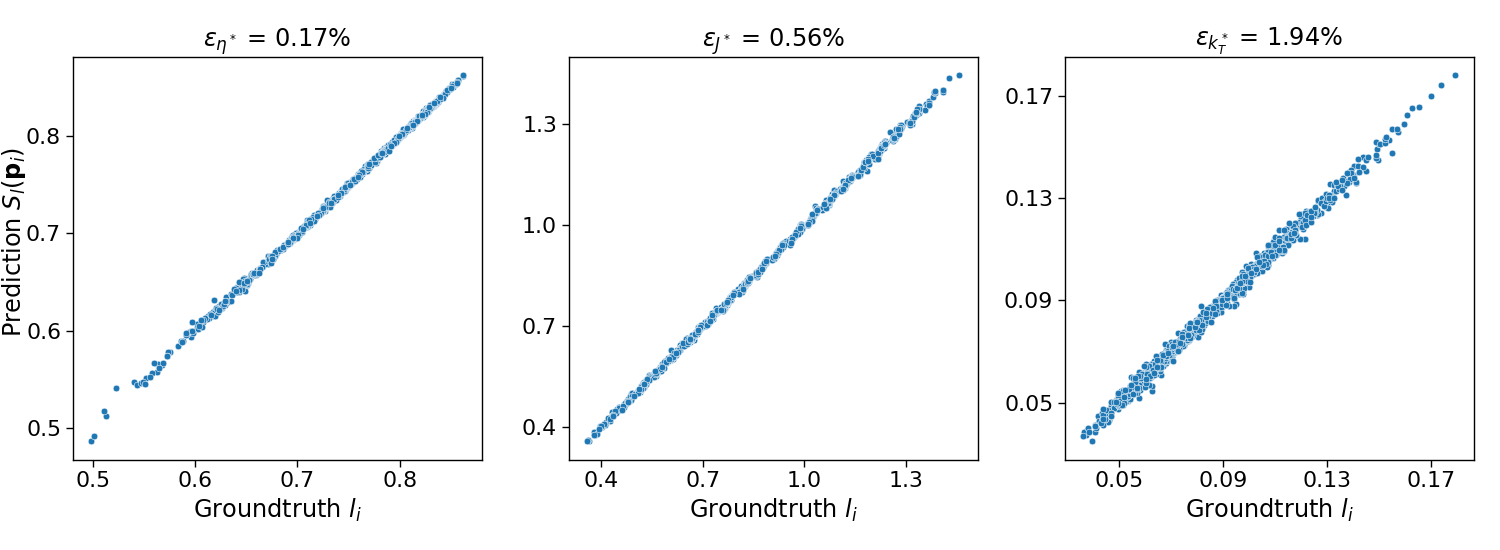}
    \caption{Scatter plot of the surrogate models comparing target and prediction values for each of the $1000$ target label vectors in the test set}
    \label{fig:scatter-su}
\end{figure}

To quickly validate different hyperparameter configurations and architectures of CFM models, surrogate models were used to approximate the flow simulation of Section \ref{chap:PropData}. This led to a significantly reduced computational cost while maintaining the accuracy within acceptable limits. For each label $l\in\{\eta^*,J^*,k_T^*\}$, a neural network $S_l^\theta\colon\R^6\to\R$ was trained to predict the label value from a given parameter vector $\p$. Aside from using $6$ layers, as well as $500$ and $250$ epochs for total training and learning rate reduction, respectively, each surrogate model used the same architecture and training parameters as the CFM model from sub-section \ref{sect:FMModel}. A simple mean squared error over training batches  was used as training loss. For validation, the Mean Relative Error (MRE,  Eq. \eqref{eq:mre}) between target label values $l_i, l\in\{\eta^*,J^*,k_T^*\}$ and surrogate model predictions $S_l^{\theta}(\p_i)$ over the test dataset $\mathcal{D}_{\mathrm{Test}}$ from Section \ref{sect:DataGenAna} was calculated.

\begin{equation}
\label{eq:mre}
    \epsilon_l^S = \frac{1}{N}\sum_{i=1}^N \left |\frac{ l_i - S_l(\p_i)}{l_i}\right |\,,\quad l\in\{\eta^*,J^*,k_T^*\}.
\end{equation}
The test dataset consists of $N=1000$ pairs of parameter and label vectors $[\p_i,\la_i]$. The error values for each label, as well as parity plots between the predictions of the surrogate model $S_l(\mathbf{p})$ and the ground truth labels $l$ for all $[\p_i,\la_i]\in\mathcal{D}_{\mathrm{Test}}$ are given in Figure \ref{fig:scatter-su}.
The high accuracy achieved by the surrogate models made it furthermore possible to study the application of synthetic data augmentation, which will be covered in sub-section \ref{sec_val_data_aug}.

\subsection{Time and Resources}
\label{sect:TimeRes}
The training and testing of CFM and surrogate models were performed on an Nvidia GeForce GTX 1050 Ti GPU with 4 GB of memory and on a computer with an AMD Ryzen 9 900x 12-core CPU and 64 GB of RAM.
With this computational setup, a CFM model was trained within $15$ minutes, while three surrogate models for the three performance labels were trained within $2$ minutes total. 
In contrast, a single CFD simulation with OpenProp was performed in approximately $10$ seconds and, consequently, the generation of the training and test set took around $6$ hours. Generally, obtaining label vectors for 1000 parameter vectors $p$ took roughly 3 hours. While introducing some minor error, surrogate models reduced this time to around $3\pm 1$ seconds. The time needed for OpenProp is admittedly rather short. Therefore, it needs to be pointed out again that the focus of this paper lies on methodology. If a RANS code was used instead, the speed-up would naturally be much higher. 

\section{Validation}
\label{chap:Results}
Four studies were performed to measure the capacity of trained CFM models to generate geometrically diverse design options of ship propellers that accurately match specified performance requirements. In sub-section \ref{sec_val_accuracy}, the accuracy of generated designs is tested for a wide variety of performance targets. The diversity of generated designs for one fixed vector of performance label values is demonstrated in sub-section \ref{sec_val_diversity}. Finally, the high accuracy of the surrogate models (see sub-section \ref{sect:SUModel}) motivated experiments on synthetic data augmentation when only a low amount of training data is available, which are covered in Section \ref{sec_val_data_aug}. Finally, the generative leaning workflow is again applied to a real-world design problem in sub-section \ref{sec:past_design_task}.

\subsection{Accuracy of Designs}\label{sec_val_accuracy}
\begin{figure}[h!]
    \centering
    \includegraphics[width=1\linewidth]{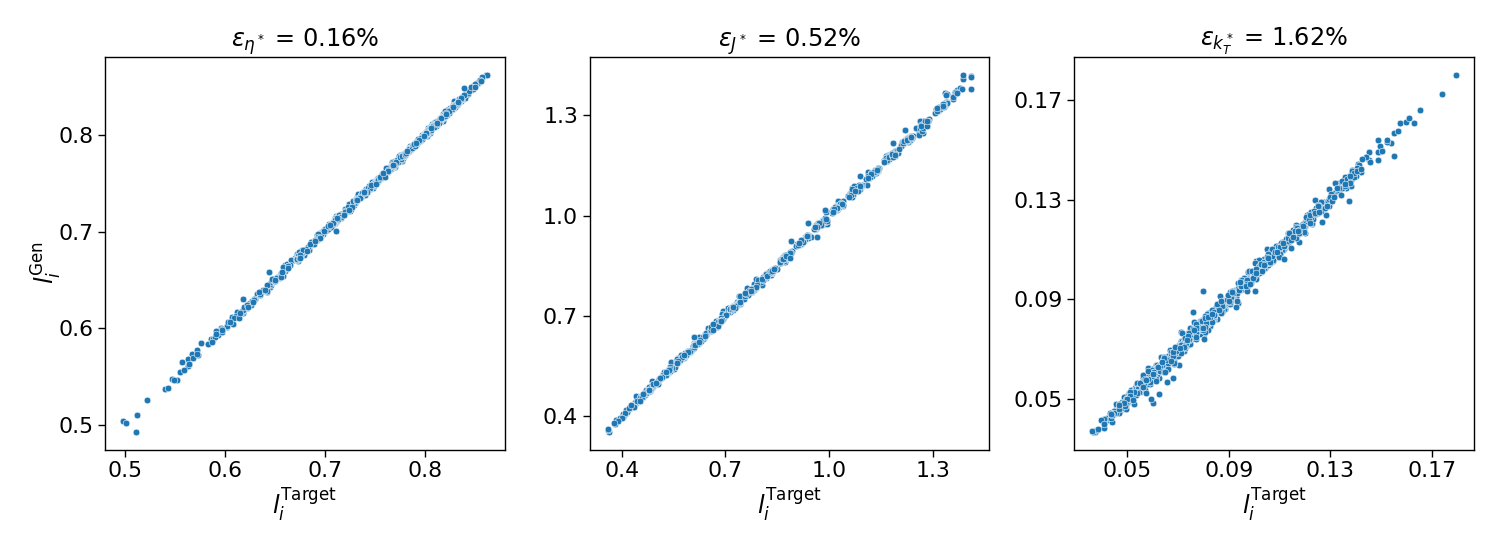}
    \caption{Comparison of target label values $l_i^\text{Target}$ (x-axis) and groundtruth label values $l_i^\text{Gen}$ (y-axis) of each label $l_i\in (\eta^*,J^*,k_T^*)$ (left/center/right) for $N=949$ valid generated designs}
    \label{fig:scatter-cfm}
\end{figure}

Considering the test dataset $\mathcal{D}_{\mathrm{Test}}=\{[\p_i,\la_i]\}_{i=1}^{1000}$ from sub-section \ref{sect:DataGenAna}, the trained CFM model was used to generate one design $\p_i^{\mathrm{Gen}}$ for each label vector $\la_i$. Afterwards, the true label vectors $\la_i^{\mathrm{Gen}}$ for the generated $\p_i^{\mathrm{Gen}}$ were obtained via the original simulation workflow (Section \ref{chap:PropData}) used to initially generate training and test datasets. Figure \ref{fig:scatter-cfm} shows parity plots between the target performance label values $\la_i$ and the true label values $l_i^{\mathrm{Gen}}$ for all labels $l\in\{\eta^*,J^*,k_T^*\}$, as well as mean relative errors (see Eq. \ref{eq:mre}) between $l_i$ and $l_i^{\mathrm{Gen}}$ for each label). 
High accuracy is achieved by the generated designs for all target values of all three performance labels. Due to the strong linear and logarithmic correlations (see Figure \ref{fig:pairplot}) between the design parameter $P$ and the labels $\eta^*$ and $J^*$, respectively, accuracy for these labels is slightly higher than for $k_T^*$.

\subsection{Diversity of Designs}\label{sec_val_diversity}
\begin{figure}
    \centering
    \includegraphics[width=1\linewidth]{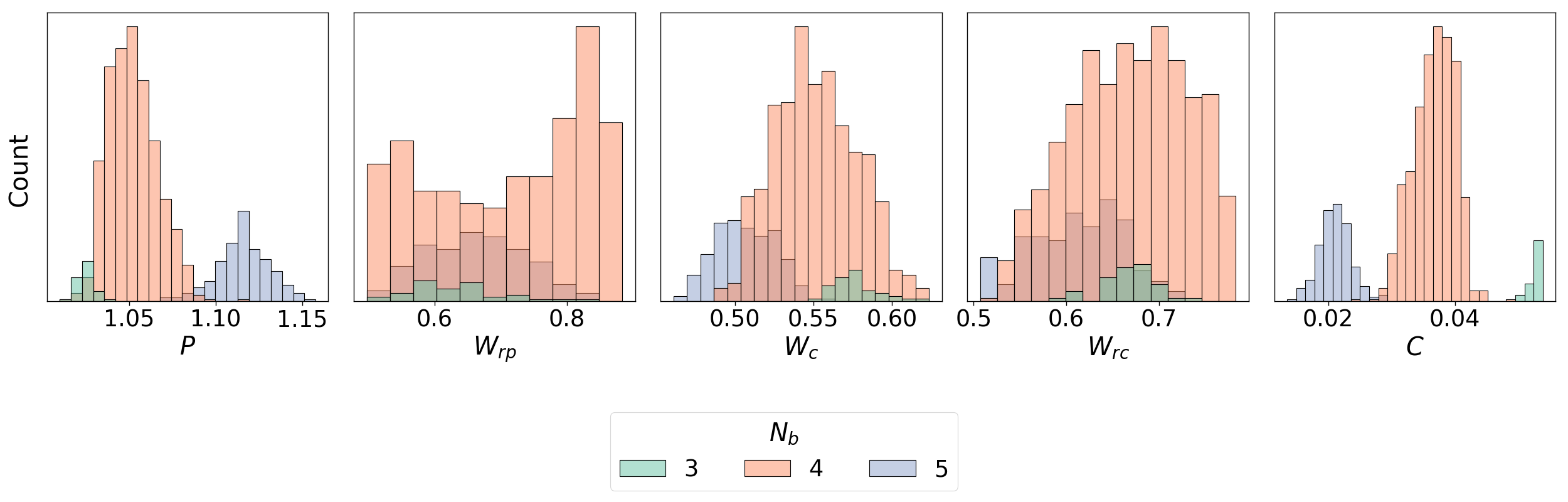}
    \caption{Distributions of the generated design parameters $p^\text{Gen}$ for $\la^\text{Target}=(0.8, 1.0, 0.1)$}
    \label{fig:hist-params}
\end{figure}

\begin{figure}
    \centering
    \includegraphics[width=\textwidth]{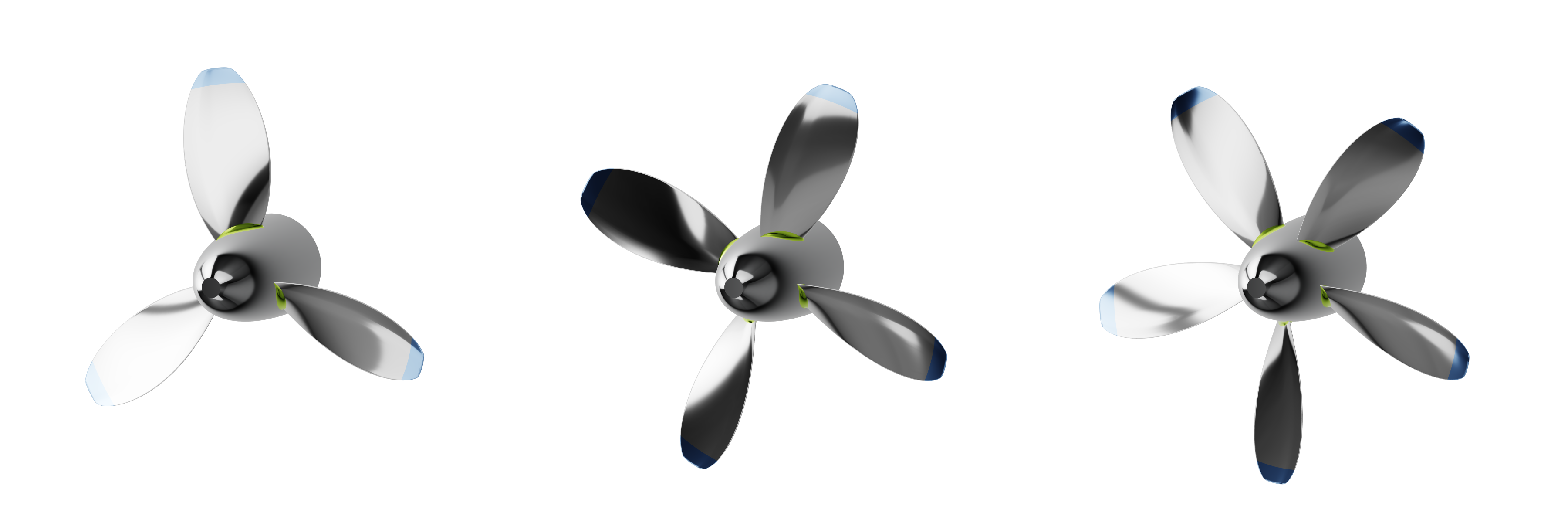}
    \caption{Three exemplary propeller designs $\p_i^{\mathrm{Gen}}$ for $\la^\text{Target}=(0.8, 1.0, 0.1)$ generated by the CFM model}
    \label{fig:gen-props}
\end{figure}

\begin{figure}
    \centering
    \includegraphics[width=1\linewidth]{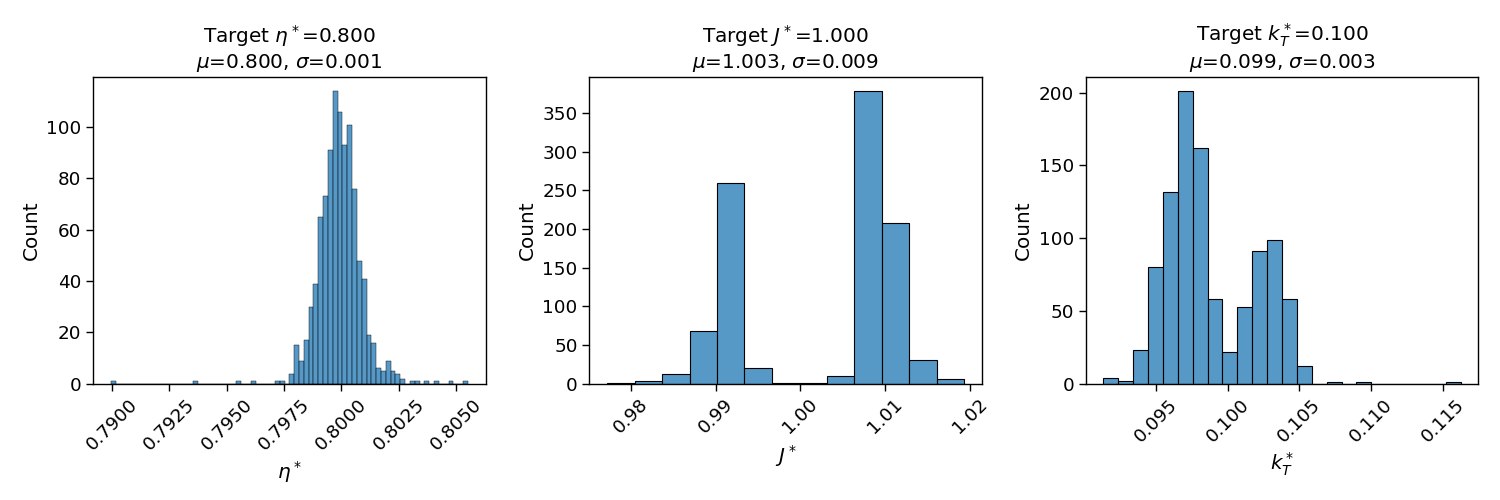}
    \caption{Distributions of the achieved labels $l_i^\text{Gen}$ by the generated designs for $\la^\text{Target}=(0.8, 1.0, 0.1)$}
    \label{fig:hist-labels}
\end{figure}

To investigate the geometrical diversity of design options generated by the CFM models, 1000 designs $\p_i^{\mathrm{Gen}}$were generated for the fixed target label vector
\begin{align}
    \la^{\mathrm{Target}}=(\eta^*,J^*,k_T^*)=(0.8,1.0,0.1).
\end{align}
The histogram plots in Figure \ref{fig:hist-params} show the distributions of the generated values for each design parameter $\p_i^{\mathrm{Gen}}$ (see Table \ref{tab:prop-params}).
It can be seen that a meaningful variety of geometrically different designs can be obtained. This is further illustrated by Figure \ref{fig:gen-props} showing models of three generated propeller designs with differing numbers of blades $N_b$. Note that no propellers could be found with just two blades that would yield the required performance labels. As in sub-section \ref{sec_val_accuracy}, the true label values $l_i^{\mathrm{Gen}}$ were obtained for all $\p_i^{\mathrm{Gen}}$ by the simulation workflow. Figure \ref{fig:hist-labels} shows the distributions of all $l_i^{\mathrm{Gen}}$ for $l\in\{\eta^*,J^*,k_T^*\}$. It can be seen that all true label values fall within close ranges of their respective target value. 


\subsection{Data Augmentation for Reduced Training Datasets}\label{sec_val_data_aug}
\begin{figure}
    \centering
    \includegraphics[width=\linewidth]{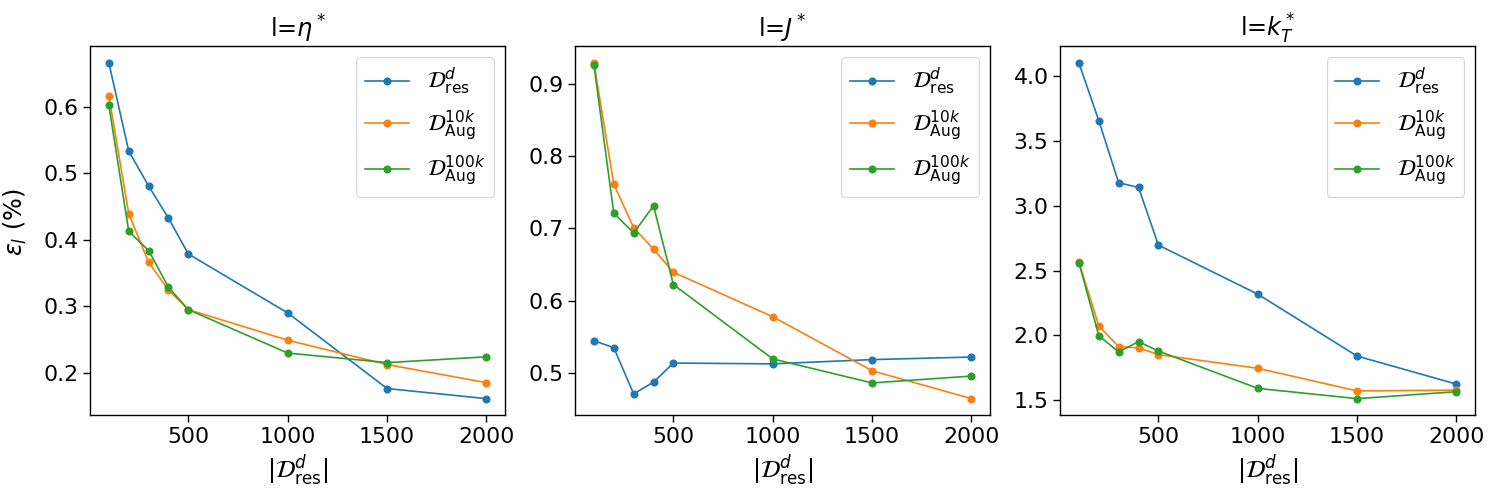}
    \caption{Plots of the MREs of the CFM models for each individual label}
    \label{fig:fm-performances}
\end{figure}

The high accuracy achieved by the surrogate models in sub-section \ref{sect:SUModel} motivated experiments on data augmentation, which describes the creation of larger, synthetic training datasets $\mathcal{D}_{\mathrm{Aug}}$. For these datasets, a large number of parameter vectors is again sampled via Latin Hypercube as in Section \ref{chap:PropData}. Instead of obtaining the corresponding label vectors through the original, time intensive simulation workflow, they are rapidly predicted by the surrogate models:
\begin{align}\label{eq:surr_pred}
    l_i^{\mathrm{Aug}}=S_l(\p_i),\quad l\in\{\eta^*,J^*,k_T^*\}.
\end{align}
This approach enables the generation of much larger training datasets in short time (see sub-section \ref{sect:TimeRes}) at the cost of introducing some error. As original training data may be far more expensive to obtain in real world use cases, (see e.g. \cite{GasTurb}), data augmentation presents an interesting approach to achieve better training results when limited data are available. To simulate such a scenario, surrogate models $S_l^d$ were trained on restricted sets $\mathcal{D}_{\mathrm{res}}^d$ of initial training data with varying sizes
\begin{align}\label{eq:Init_data_sizes}
    \lvert\mathcal{D}_{\mathrm{res}}^d\rvert=d\in\{100,200,300,400,500,1000,1500,2000\}.
\end{align}
All $\mathcal{D}_{\mathrm{res}}^d$ were sampled from the original dataset $\mathcal{D}$, hence $\mathcal{D}_{\mathrm{res}}^{2000}=\mathcal{D}$. For each $d$ listed in Eq. \eqref{eq:Init_data_sizes}, the surrogate models $S_l^d$ trained on $\mathcal{D}_{\mathrm{res}}^d$ were then used as described above and in Eq. \eqref{eq:surr_pred} to generate synthetic datasets $\mathcal{D}_{\mathrm{Aug}}^{10k}$ and $\mathcal{D}_{\mathrm{Aug}}^{100k}$ containing 10\,000 and 100\,000 pairs $[\p_i,\la_i^{\mathrm{Aug}}]$, respectively. Three CFM models were then trained on $\mathcal{D}_{\mathrm{Aug}}^{10k}$, $\mathcal{D}_{\mathrm{Aug}}^{100k}$ and $\mathcal{D}_{\mathrm{res}}^d$. Performance of each model evaluated on the test Dataset $\mathcal{D}_{\mathrm{Test}}$ as in sub-section \ref{sec_val_accuracy}. For each label $l\in\{\eta^*,J^*,k_T^*\}$ and each initial dataset size $d$, the MRE (Eq. \ref{eq:mre}) between target labels true labels are plotted for $\mathcal{D}_{\mathrm{Aug}}^{10k}$, $\mathcal{D}_{\mathrm{Aug}}^{100k}$ and $\mathcal{D}_{\mathrm{res}}^d$ in Figure \ref{fig:fm-performances}. The corresponding numerical values are contained in Table \ref{tab:fm-performances-tab}. As before, all true label values $l^{\mathrm{Gen}}$ were obtained by the workflow from Section \ref{chap:PropData}.

It can be observed that the positive effect of data augmentation increases with the complexity of the considered label. $J^*$ is the least complex label due to the strong linear correlation with the parameter $P$. With $J^*$, CFM Models trained on augmented datasets are outperformed by the baseline model trained on $\mathcal{D}_{\mathrm{res}}^d$ for all $d\leq 1000$ and yield only a slight improvement for $d=1500$ and $d=2000$. $\eta^*$ is more complex than $J^*$, however there still is a strong logarithmic correlation with $P$. Here, significant improvements in accuracy can be observed for all $d\leq 1000$. For $d\geq 1000$, augmented models are outperformed slightly by the baseline model. The most complex label is $k_T^*$, for which no strong correlations with the design parameters can be observed in Figure \ref{fig:pairplot}. For this label, augmented models outperform the baseline trained on $\mathcal{D}_{\mathrm{res}}^d$ on all $d$ considered in this experiment, yielding a the greatest increase in accuracy over all considered labels. This improvement is most significant for lower initial dataset sizes $d$ and decreases towards $d=2000$.

\begin{table}
    \tbl{Relative improvements (in \%) in MRE of models trained on augmented datasets compared to $\mathcal{D}_{\mathrm{res}}^d$ for $l\in\{\eta^*,J^*,k_T^*\}$ Negative values denote a reduction in error.}{
    \begin{tabular}{c c c c c c c c c c}
        \toprule
        & \multicolumn{2}{c}{$\eta^*$} & \multicolumn{2}{c}{$J^*$} & \multicolumn{2}{c}{$k_T^*$} \\
        \cmidrule(lr){2-3} \cmidrule(lr){4-5} \cmidrule(lr){6-7}
        $\lvert\mathcal{D}_{\mathrm{res}}^d\rvert$ & 
        $\mathcal{D}_{\mathrm{Aug}}^{10k}$ & $\mathcal{D}_{\mathrm{Aug}}^{100k}$ & $\mathcal{D}_{\mathrm{Aug}}^{10k}$ & $\mathcal{D}_{\mathrm{Aug}}^{100k}$ & $\mathcal{D}_{\mathrm{Aug}}^{10k}$ & $\mathcal{D}_{\mathrm{Aug}}^{100k}$\\
        \midrule
        
        100 & -7.45 & -9.57 & 70.51 & 70.05 & -37.5 & -37.59 \\
        200 & -17.69 & -22.62 & 42.25 & 34.75 & -43.35 & -45.41 \\
        300 & -23.85 & -20.19 & 48.78 & 47.23 & -39.87 & -41.15 \\
        400 & -25.01 & -24.02 & 37.72 & 50.03 & -39.43 & -37.9 \\
        500 & -22.11 & -22.12 & 24.44 & 21.12 & -31.37 & -30.35 \\
        1000 & -14.14 & -20.82 & 12.7 & 1.33 & -24.71 & -31.41 \\
        1500 & 20.5 & 22.11 & -2.97 & -6.22 & -14.63 & -17.87 \\
        2000 & 14.9 & 39.08 & -11.02 & -5.04 & -2.85 & -3.6 \\
        \bottomrule
    \end{tabular}}
    
    \label{tab:fm-performances-tab}
\end{table}

A possible explanation for the observations made is that labels with strong correlations are easier to learn by the CFM models, even when low amounts of original data are available. Thus, the error introduced by the surrogate models might outweigh the benefits of a larger training dataset. Overall, the results presented here indicate that data augmentation via surrogate models might be a viable approach to increase model performance, especially when only low amounts of training data describing complex parameter-label relationships are given. 
\subsection{Application to a Design Task}\label{sec:past_design_task}
\begin{figure}
    \centering
    \includegraphics[width=0.7\textwidth]{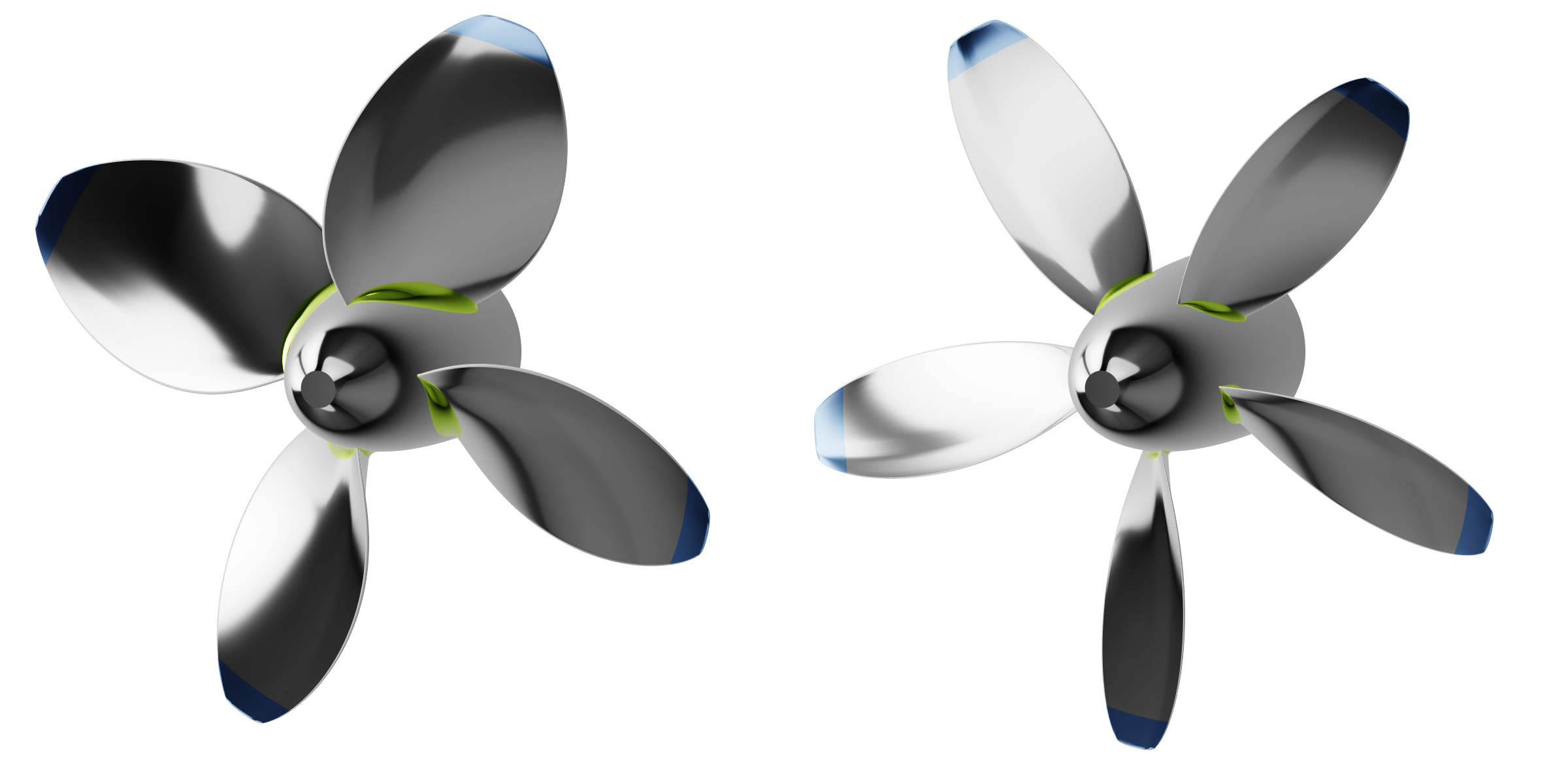}
    \caption{Two propeller designs generated by the CFM model}
    \label{fig:designTask-props}
\end{figure}

Finally, the proposed workflow was applied to a past design task for the propeller of a lightweight motorboat. At the time, the objective of the design task was to develop a propeller to reach the maximal possible ship speed under the given constraints - the engine rpm, the maximum torque, the maximal propeller diameter and the boat's resistance over speed.\\
The parametric propeller model used in the original optimization was also set up in CAESES but its functionality, i.e., the design variables affecting the radial distributions, was more comprehensive and for the simulations a RANS solver was used. For these reasons, no direct comparison between the original optimized design and the newly generated designs will be made.\\
To demonstrate one possible workflow, the new design task is now to generate a variety of propellers for one ship speed and the given engine rpm and propeller diameter. This allows a propeller designer to choose between different suitable designs based on manufacturing constraints or other performance constraints, e.g., cavitation risk.\\
The targeted ship speed determines the required thrust which gives the performance label $k_T$, here $k_T = 0.1402$, and the label $J$ is calculated with $J=\frac{V_A}{nD} = 1.209$. The third label $\eta$ was not fixed, i.e., the generative model provided designs in the range determined by the propeller dataset used for model training.\\
100 designs were generated but only seven designs were valid after validating them using the surrogate models, i.e., the targeted performance labels were not reached by the generated designs within a tolerance. An explanation for this lower number of valid designs could be the free label $\eta$,  as other studies with a fixed $\eta$ have shown valid designs rates of up to 100\%.\\
The seven valid designs were then evaluated for the torque constraint, which all of them complied with. Two of these designs are shown in \autoref{fig:designTask-props}.\\

The two designs were additionally validated using the OpenProp simulations and show good agreement with the target performance labels, see \autoref{tab:applicationValidation}.

\begin{table}[h]
  \tbl{Validation of the generated propeller design}
  {\begin{tabular}{clcc}
    \toprule
    \textbf{} & \multicolumn{1}{c}{\textbf{$\eta$}} & \textbf{$J$} & \textbf{$k_T$} \\ \midrule
    \textbf{Design with 4 blades} &  & &\\
    target & 0.7847 & 1.209 & 0.1402\\
    surrogate prediction & 0.7856 & 1.2138 & 0.1391\\
    OpenProp simulation & 0.7827 & 1.214 & 0.141\\ 
    \midrule
    \textbf{Design with 5 blades} &  & &\\
    target & 0.8082 & 1.209 & 0.1402\\
    surrogate prediction & 0.8081 & 1.2172 & 0.1393\\
    OpenProp simulation & 0.8078 & 1.21 & 0.1422\\ \bottomrule
  \end{tabular}}
  \label{tab:applicationValidation}
\end{table}

\section{Conclusion and Future Work}
\label{chap:Summary}
This article examined the application of generative Conditional Flow Matching models for the inverse design of ship propellers. The training and the test set for the models consist of six design variables that control a fully-parametric propeller model in CAESES and three performance labels derived from a simulations using OpenProp. In addition, the use of surrogate models of flow simulations with efficient predictions of performance labels were successfully applied for rapid testing of model configurations, as well as for data augmentation. Validation studies were performed on the CFM models, proving the capability of producing diverse design alternatives that accurately meet a wide variety of performance requirements. Lastly, the potential benefits of synthetic data augmentation via surrogate models were shown for use cases where training data is of low availability.  
\\
Building on the results of this article, subsequent studies may investigate the performance of CFM models when increasing the number of propeller parameters and its performance labels, where the inherent relationship might be more complex. For example,  the skew and rake of a propeller and a cavitation criterion in parameters and labels, respectively. In this context, one may also use another more sophisticated CFD method that takes these new parameters and labels into account. To be of practical use in the future, also the interaction of the hull and the propeller would have to be considered. The higher complexity of these realistic scenarios will require considerably more computational resources. It underlines that in engineering data are scarce, calling for efficient methods of generative design.

\section{Acknowledgment}
Parts of this work have been undertaken within the R\&D-project VIT-VI (short for German: “ Virtuelle Triebwerksentwicklung mit Verfahren der künstlichen Intelligenz” | English: “Virtual Engine Development using artificial intelligence methods”), project number 85063176. The project is funded by the program for the promotion of research, innovations, and technologies of the state of Brandenburg (ProFIT Brandenburg) through the Ministry of Economic Affairs, Labor, and Energy of the state of Brandenburg (MWAE) and supported by the Investment Bank of the state of Brandenburg (ILB). The funding comes from the European Regional Development Fund (EFRE) and the state of Brandenburg. The VITVI research project is co-funded by the European Union.\footnote{VIT-VI Project: \url{https://www.friendship-systems.com/company/r_a_d/vit-vi/}}. Hanno Gottschalk also acknowledges financial support by the Germen research council through SPP2403 "Carnot Batteries" project GO 833/8-1 "Inverse aerodynamic design of turbo components for Carnot batteries by means of physics
informed networks enhanced by generative learning".
\section{Use of Generative AI}
ChatGPT by OpenAI\footnote{OpenAI Homepage: https://openai.com}, versions 4o, 5 and 5.1, was occasionally used to aid in writing by providing better formulations for self-written paragraphs.  
\bibliographystyle{tfcad}
\bibliography{interactcadsample}

@article{GasTurb,
    title = {Generative Design of a Gas Turbine Combustor Using Invertible Neural Networks},
    author = {Krueger, Patrick and Gottschalk, Hanno and Werdelmann, Bastian and Krebs, Werner},
    journal = {Journal of Engineering for Gas Turbines and Power},
    volume = {147},
    number = {1},
    year = {2025},
    publisher = {American Society of Mechanical Engineers Digital Collection}
}

@inproceedings{chan2023lu,
  title={Lu-net: Invertible neural networks based on matrix factorization},
  author={Chan, Robin and Penquitt, Sarina and Gottschalk, Hanno},
  booktitle={2023 International Joint Conference on Neural Networks (IJCNN)},
  pages={1--10},
  year={2023},
  organization={IEEE}
}

@software{DallE,
    author = {OpenAI},
    title = {DALL-E 3},
    year = {2021},
    url = {https://openai.com/index/dall-e-3/},
    urldate = {2025-06-20}
}

@software{ChatGPT,
    author = {OpenAI},
    title = {ChatGPT},
    year = {2022},
    url = {https://chatgpt.com/},
    urldate = {2025-06-20}
}

@misc{GAN,
    title={Generative Adversarial Networks}, 
    author={Ian J. Goodfellow and Jean Pouget-Abadie and Mehdi Mirza and Bing Xu and David Warde-Farley and Sherjil Ozair and Aaron Courville and Yoshua Bengio},
    year={2014},
    eprint={1406.2661},
    archivePrefix={arXiv},
    primaryClass={stat.ML},
    url={https://arxiv.org/abs/1406.2661}, 
}

@misc{VAE,
    title={Auto-Encoding Variational Bayes}, 
    author={Diederik P Kingma and Max Welling},
    year={2022},
    eprint={1312.6114},
    archivePrefix={arXiv},
    primaryClass={stat.ML},
    url={https://arxiv.org/abs/1312.6114}, 
}

@misc{RL,
    title={Reinforcement Learning: A Survey}, 
    author={L. P. Kaelbling and M. L. Littman and A. W. Moore},
    year={1996},
    eprint={cs/9605103},
    archivePrefix={arXiv},
    primaryClass={cs.AI},
    url={https://arxiv.org/abs/cs/9605103}, 
}

@Inbook{GMM,
    author="Reynolds, Douglas",
    editor="Li, Stan Z.
    and Jain, Anil",
    title="Gaussian Mixture Models",
    bookTitle="Encyclopedia of Biometrics",
    year="2009",
    publisher="Springer US",
    address="Boston, MA",
    pages="659--663",
    isbn="978-0-387-73003-5",
    doi="10.1007/978-0-387-73003-5_196",
    url="https://doi.org/10.1007/978-0-387-73003-5_196"
}

@misc{InvProbINN,
    title={Analyzing Inverse Problems with Invertible Neural Networks}, 
    author={Lynton Ardizzone and Jakob Kruse and Sebastian Wirkert and Daniel Rahner and Eric W. Pellegrini and Ralf S. Klessen and Lena Maier-Hein and Carsten Rother and Ullrich Köthe},
    year={2019},
    eprint={1808.04730},
    archivePrefix={arXiv},
    primaryClass={cs.LG},
    url={https://arxiv.org/abs/1808.04730}, 
}

@misc{AffCoupling,
    title={Density estimation using Real NVP}, 
    author={Laurent Dinh and Jascha Sohl-Dickstein and Samy Bengio},
    year={2017},
    eprint={1605.08803},
    archivePrefix={arXiv},
    primaryClass={cs.LG},
    url={https://arxiv.org/abs/1605.08803}, 
}

@misc{GenAIShipDesign,
    title={Generative AI in Ship Design}, 
    author={Sahil Thakur and Navneet V Saxena and Prof Sitikantha Roy},
    year={2024},
    eprint={2408.16798},
    archivePrefix={arXiv},
    primaryClass={cs.LG},
    url={https://arxiv.org/abs/2408.16798}, 
}

@inproceedings{AirfoilCGAN,
    author={Tan, Xavier and Manna, Dai and Chattoraj, Joyjit and Yong, Liu and Xinxing, Xu and Ha, Dao My and Feng, Yang},
    booktitle={2022 17th International Conference on Control, Automation, Robotics and Vision (ICARCV)}, 
    title={Airfoil Inverse Design using Conditional Generative Adversarial Networks}, 
    year={2022},
    pages={143-148},
    doi={10.1109/ICARCV57592.2022.10004343}
}

@article{AirfoilVAE,
    title={Inverse design optimization framework via a two-step deep learning approach: application to a wind turbine airfoil},
    volume={39},
    ISSN={1435-5663},
    url={http://dx.doi.org/10.1007/s00366-022-01617-6},
    DOI={10.1007/s00366-022-01617-6},
    number={3},
    journal={Engineering with Computers},
    publisher={Springer Science and Business Media LLC},
    author={Yang, Sunwoong and Lee, Sanga and Yee, Kwanjung},
    year={2022},
    month=mar, pages={2239–2255} 
}

@article{InvDesignMaterial,
    title = {Machine Learning-Based Methods for Materials Inverse Design: A Review},
    journal = {Computers, Materials and Continua},
    volume = {82},
    number = {2},
    pages = {1463-1492},
    year = {2025},
    issn = {1546-2218},
    doi = {https://doi.org/10.32604/cmc.2025.060109},
    url = {https://www.sciencedirect.com/science/article/pii/S1546221825001626},
    author = {Yingli Liu and Yuting Cui and Haihe Zhou and Sheng Lei and Haibin Yuan and Tao Shen and Jiancheng Yin}
}

@misc{GenAIEngineering,
      title={Deep Generative Models in Engineering Design: A Review}, 
      author={Lyle Regenwetter and Amin Heyrani Nobari and Faez Ahmed},
      year={2022},
      eprint={2110.10863},
      archivePrefix={arXiv},
      primaryClass={cs.LG},
      url={https://arxiv.org/abs/2110.10863}, 
}

@book{PropTheory,
	title={Theory of Wing Sections, Including a Summary of Airfoil Data},
	author={Abbott, I.H. and Von Doenhoff, A.E.},
	isbn={9781306346818},
	series={Dover Books on Aeronautical Engineering Series},
	url={https://books.google.de/books?id=lWe8AQAAQBAJ},
	year={2012},
	publisher={Dover Publications}
}

@book{Naca66,
	title={Minimum Pressure Envelopes for Modified NACA-66 Sections with NACA A=0.8 Camber and Buships Type 1 and Type 2 Sections},
	author={Brockett, Terry},
        pages={42},
	url={https://apps.dtic.mil/sti/tr/pdf/AD0629379.pdf},
	year={1966},
	publisher={Defense Technical Information Center}
}

@book{VLM,
	place={Cambridge}, 
	edition={2}, 
	series={Cambridge Aerospace Series}, 
	title={Low-Speed Aerodynamics}, 
	publisher={Cambridge University Press}, 
	author={Katz, Joseph and Plotkin, Allen}, 
	year={2001},
	isbn={978-0-521-66219-2}
}

@inproceedings{OpenProp,
  title={OpenProp: An Open-Source Design Tool for Propellers and Turbines},
  author={Epps, B. P. and Stanway, M. J. and Kimball, R. W.},
  booktitle={SNAME Propeller and Shafting Symposium},
  pages={D021S002R003},
  year={2009},
  organization={SNAME}
}

@article{LHS,
    author = {Mckay, M. and Beckman, Richard and Conover, William},
    year = {1979},
    month = {05},
    pages = {239-245},
    title = {A Comparison of Three Methods for Selecting Vales of Input Variables in the Analysis of Output From a Computer Code},
    volume = {21},
    journal = {Technometrics},
    doi = {10.1080/00401706.1979.10489755}
}

@misc{FlowMatchLipman,
	title={Flow Matching for Generative Modeling}, 
	author={Yaron Lipman and Ricky T. Q. Chen and Heli Ben-Hamu and Maximilian Nickel and Matt Le},
	year={2023},
	eprint={2210.02747},
	archivePrefix={arXiv},
	primaryClass={cs.LG},
	url={https://arxiv.org/abs/2210.02747}, 
}

@misc{CondFlowMatchTong,
	title={Improving and generalizing flow-based generative models with minibatch optimal transport}, 
	author={Alexander Tong and Kilian Fatras and Nikolay Malkin and Guillaume Huguet and Yanlei Zhang and Jarrid Rector-Brooks and Guy Wolf and Yoshua Bengio},
	year={2024},
	eprint={2302.00482},
	archivePrefix={arXiv},
	primaryClass={cs.LG},
	url={https://arxiv.org/abs/2302.00482}, 
}

@misc{NeuralODEs,
	title={Neural Ordinary Differential Equations}, 
	author={Ricky T. Q. Chen and Yulia Rubanova and Jesse Bettencourt and David Duvenaud},
	year={2019},
	eprint={1806.07366},
	archivePrefix={arXiv},
	primaryClass={cs.LG},
	url={https://arxiv.org/abs/1806.07366}, 
}

@software{TorchCFM,
    author = {Alexander Tong and Kilian Fatras and Nikolay Malkin and Guillaume Huguet and Yanlei Zhang and Jarrid Rector-Brooks and Guy Wolf and Yoshua Bengio},
    title = {TorchCFM: a Conditional Flow Matching library},
    year = {2023},
    url = {https://github.com/atong01/conditional-flow-matching},
    urldate = {2025-06-20}
}

@article{fukushima1975cognitron,
  title={Cognitron: A self-organizing multilayered neural network},
  author={Fukushima, Kunihiko},
  journal={Biological cybernetics},
  volume={20},
  number={3},
  pages={121--136},
  year={1975},
  publisher={Springer}
}

@article{ardizzone2018analyzing,
  title={Analyzing inverse problems with invertible neural networks},
  author={Ardizzone, Lynton and Kruse, Jakob and Wirkert, Sebastian and Rahner, Daniel and Pellegrini, Eric W and Klessen, Ralf S and Maier-Hein, Lena and Rother, Carsten and K{\"o}the, Ullrich},
  journal={arXiv preprint arXiv:1808.04730},
  year={2018}
}

@article{kingma2018glow,
  title={Glow: Generative flow with invertible 1x1 convolutions},
  author={Kingma, Durk P and Dhariwal, Prafulla},
  journal={Advances in neural information processing systems},
  volume={31},
  year={2018}
}

@article{jia2025inverse,
  title={Inverse Design of Compressor/Fan Blade Profiles Based on Conditional Invertible Neural Networks},
  author={Jia, Xinkai and Huang, Xiuquan and Jiang, Shouyong and Wang, Dingxi and Firrone, Christian Maria},
  journal={Journal of Turbomachinery},
  volume={147},
  number={10},
  pages={101014},
  year={2025},
  publisher={American Society of Mechanical Engineers}
}

@article{yang2023normalizing,
  title={Normalizing flows for efficient inverse design of thermophotovoltaic emitters},
  author={Yang, Jia-Qi and Xu, YuCheng and Fan, Kebin and Wu, Jingbo and Zhang, Caihong and Zhan, De-Chuan and Jin, Biao-Bing and Padilla, Willie J},
  journal={ACS Photonics},
  volume={10},
  number={4},
  pages={1001--1011},
  year={2023},
  publisher={ACS Publications}
}

@inproceedings{rezende2015variational,
  title={Variational inference with normalizing flows},
  author={Rezende, Danilo and Mohamed, Shakir},
  booktitle={International conference on machine learning},
  pages={1530--1538},
  year={2015},
  organization={PMLR}
}
\end{document}